\documentclass{article}

\usepackage{microtype}
\usepackage{graphicx}
\usepackage{booktabs} 

\usepackage{hyperref}

\usepackage{amsfonts}
\usepackage{amsmath}
\usepackage{subcaption}
\usepackage{algorithm}
\usepackage{algorithmic}
\usepackage{mathtools}
\usepackage{bm}
\usepackage{booktabs} 

\usepackage{enumitem}

\usepackage[accepted]{icml2019}

\icmltitlerunning{Approximated Oracle Filter Pruning for Destructive CNN Width Optimization}

\begin{document}

\twocolumn[
\icmltitle{Approximated Oracle Filter Pruning for Destructive CNN Width Optimization}



\icmlsetsymbol{equal}{*}

\begin{icmlauthorlist}
\icmlauthor{Xiaohan Ding}{aaa}
\icmlauthor{Guiguang Ding}{aaa}
\icmlauthor{Yuchen Guo}{aaa}
\icmlauthor{Jungong Han}{bbb}
\icmlauthor{Chenggang Yan}{ccc}
\end{icmlauthorlist}

\icmlaffiliation{aaa}{Beijing National Research Center for Information Science and Technology (BNRist); School of Software, Tsinghua University, Beijing, China. Email: dxh17@mails.tsinghua.edu.cn.}
\icmlaffiliation{bbb}{WMG Data Science, University of Warwick, Coventry, United Kingdom}
\icmlaffiliation{ccc}{Institute of Information and Control, Hangzhou Dianzi University, Hangzhou, China}

\icmlcorrespondingauthor{Yuchen Guo}{yuchen.w.guo@gmail.com}
\icmlcorrespondingauthor{Jungong Han}{jungonghan77@gmail.com}

\icmlkeywords{Convolutional Neural Network, CNN, Filter Pruning, Model Compression, Model Acceleration, Efficient Deep Learning, Network Slimming}

\vskip 0.3in
]



\printAffiliationsAndNotice{}  

\begin{abstract}
It is not easy to design and run Convolutional Neural Networks (CNNs) due to: 1) finding the optimal number of filters (i.e., the width) at each layer is tricky, given an architecture; and 2) the computational intensity of CNNs impedes the deployment on computationally limited devices. Oracle Pruning is designed to remove the unimportant filters from a well-trained CNN, which estimates the filters' importance by ablating them in turn and evaluating the model, thus delivers high accuracy but suffers from intolerable time complexity, and requires a given resulting width but cannot automatically find it. To address these problems, we propose Approximated Oracle Filter Pruning (AOFP), which keeps searching for the least important filters in a binary search manner, makes pruning attempts by masking out filters randomly, accumulates the resulting errors, and finetunes the model via a multi-path framework. As AOFP enables simultaneous pruning on multiple layers, we can prune an existing very deep CNN with acceptable time cost, negligible accuracy drop, and no heuristic knowledge, or re-design a model which exerts higher accuracy and faster inference.
\end{abstract}

\section{Introduction}
Convolutional Neural Networks (CNNs) have become an important tool for many real-world applications and related research areas \cite{collobert2008unified,lecun1990handwritten,lecun1995convolutional}. Nowadays, designing a CNN usually means a tiring exploration in a vast design space, which usually includes the usage of non-linearities (ReLU, sigmoid or none), downsampling (max / average pooling or stride-2 convolution), shortcut connections \cite{he2016deep}, etc. With so many hyper-parameters in consideration, we still have to make a hard decision every time we use a convolutional layer: the number of filters, i.e., the width of the layer. Since an unnecessarily wide conv layer usually leads to meaningless parameters, heavy computational burdens, and overfitting, we wish to set a proper width for each layer, which is inherently tricky. In modern CNN architectures, some practical guidelines on the number of filters are followed. Taking VGG \cite{simonyan2014very} for example, when the feature maps are spatially downsampled by $2\times$, the number of filters becomes $2\times$, so that the computational burdens of each layer are kept roughly the same. Apparently, such guidelines leave much room to improve on the layer width for better accuracy and efficiency.

In this paper, destructive CNN width optimization refers to the process which takes a well-trained tidy CNN as input and produces an optimized one where some useless filters are removed. In this context, our method can be categorized into filter pruning, a.k.a. channel pruning \cite{he2017channel} or network slimming \cite{liu2017learning}, a family of CNN compression techniques, which features three strengths. \textbf{1) Universality:} filter pruning can handle any kinds of CNNs, making no assumptions on the application field, the network architecture or the deployment platform. \textbf{2) Effectiveness:} filter pruning effectively reduces the floating-point operations (FLOPs) of the network, which serve as the main criterion of computational burdens. When a filter is pruned, its output channel and the corresponding input channels of the following layer are removed. That is, when several conv layers stacked together are pruned respectively, the total FLOPs are reduced quadratically. \textbf{3) Orthogonality:} filter pruning simply produces a thinner network with no customized structure or extra operation, which is orthogonal to other model compression and acceleration techniques.

A common paradigm of filter pruning is to evaluate the importance of filters by some means \cite{polyak2015channel,hu2016network,li2016pruning,molchanov2016pruning,abbasi2017structural,anwar2017structured,yu2018nisp}, such that the accuracy is not damaged severely by the removal of the less important filters and can be recovered by finetuning. Apparently, the quality of the filter importance evaluation plays a vital role in the entire pipeline. By recognizing the unimportant filters, we diminish the accuracy drop, such that it becomes easier for the finetuning process to restore the accuracy. In this sense, the importance of filter can be defined by the network's accuracy drop with the filter ablated. If we ablate a filter (i.e., mask out its outputs), test the model on an assessment dataset, and observe a severe accuracy drop, and then the filter can be defined as important. 

The most straightforward and accurate algorithm to prune filters greedily by importance, which is referred to as \textit{Oracle Pruning} \cite{molchanov2016pruning}, can be implemented in a trial-and-error manner. For a specific layer, we ablate a filter, test the model on the assessment dataset, record the accuracy drop as the importance score, i.e., mean accuracy reduction \cite{abbasi2017structural} or loss increase \cite{molchanov2016pruning}, then restore the filter and move on to the next filter. When all the filters have been tested, we prune the filter with the least importance score. However, when we start to prune the next filter, the relative importance of the remaining filters may have been changed (Sect. \ref{sec-exp-1}), so they should be tested in the same manner again. In this way, we can slim a layer by always removing the filter with the least importance score until we are satisfied with the trade-off between accuracy and efficiency. However, for today's CNNs where a conv layer can comprise hundreds of filters, the time complexity of Oracle Pruning is intolerable. In order to acquire the importance score of filters with reasonable time cost, some heuristic methods \cite{li2016pruning,hu2016network,molchanov2016pruning} have been proposed, which suffer from inferior quality of importance estimation, compared to Oracle Pruning (Fig. \ref{fig-compre-pruning}). Of note is that ``oracle'' described here is only the most accurate \textit{greedy} pruning method. A better oracle would consider all combinations of pruned filters, but of course, this is extremely expensive.
\begin{figure}
	\begin{center}
		\includegraphics[width=0.9\columnwidth]{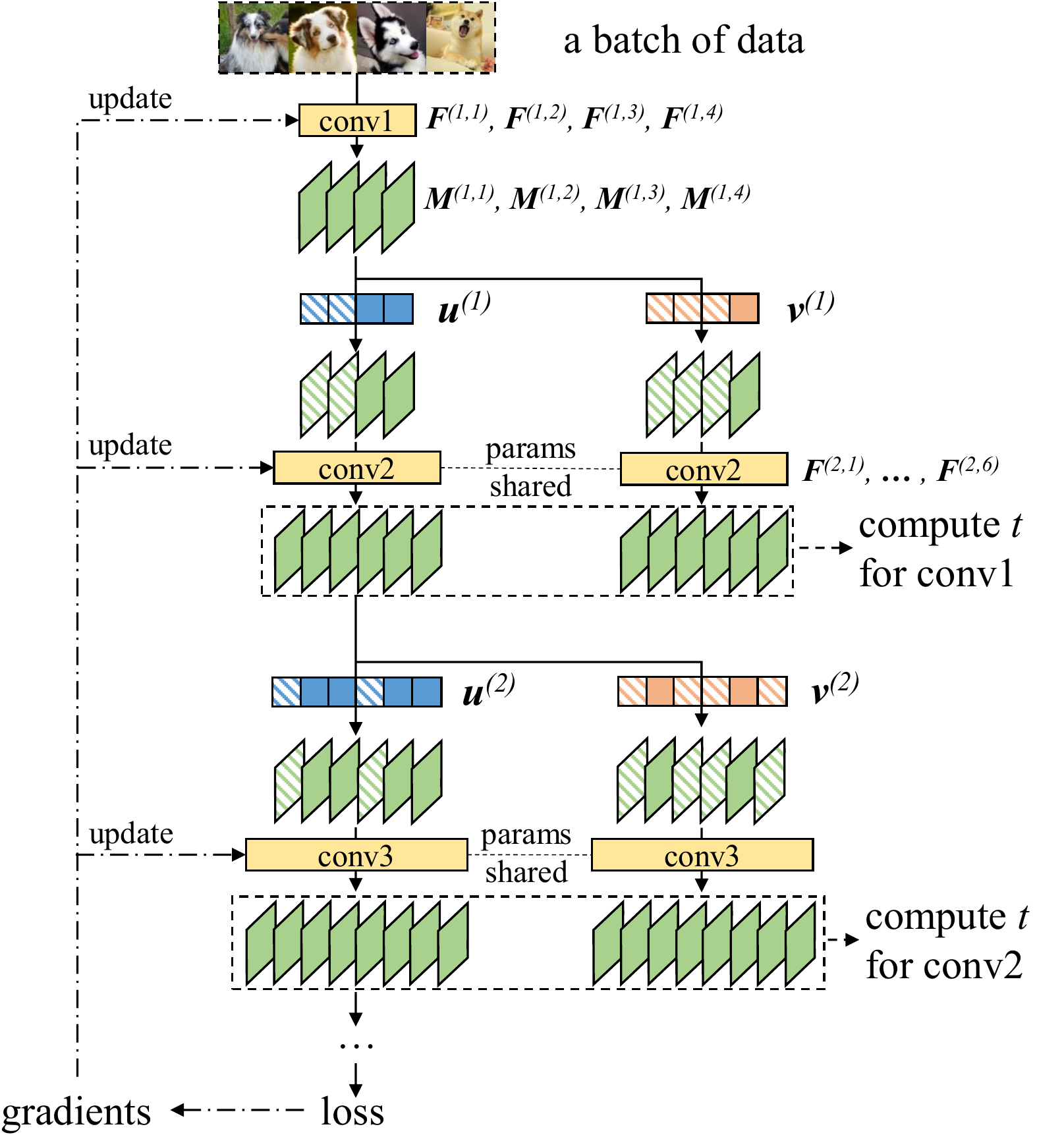}
		\caption{Overview of AOFP, where conv1 and conv2 in a CNN are being pruned simultaneously for example. Filters $\bm{F}^{(1,1)},\bm{F}^{(1,2)},\bm{F}^{(2,1)},\bm{F}^{(2,4)}$ have already been masked out, and the algorithm is trying to pick the next unimportant one out of $\{\bm{F}^{(1,3)},\bm{F}^{(1,4)}\}$ and two out of $\{\bm{F}^{(2,2)},\bm{F}^{(2,3)},\bm{F}^{(2,5)},\bm{F}^{(2,6)}\}$.}
		\label{framework}
	\end{center}
\end{figure}

In this paper, we propose Approximated Oracle Filter Pruning (AOFP), a multi-path training-time filter pruning framework (Fig. \ref{framework}), where we keep searching for the next filters to prune in a binary search manner and finetuning the model in the meantime, which features high quality of importance estimation, reasonable time complexity and no need for heuristic knowledge. The codes are available at \url{https://github.com/ShawnDing1994/AOFP}. Our contributions are summarized as follows.
\begin{itemize}[noitemsep,nolistsep,,topsep=0pt,parsep=0pt,partopsep=0pt]
	\item We improve unimportant filters selection by analyzing the outputs of the next layer only, rather than the final outputs. However, instead of solving a linear regression problem layer-by-layer \cite{luo2017thinet,luo2018thinet}, we ablate the filters randomly, then compute and accumulate the change in the next layer's outputs, which is referred to as Damage Isolation. Doing so enables not only the faster importance estimation but also mutually independent pruning on all the layers simultaneously. We have shown that the structural change of every layer in CNNs can be separately measured using only local information, which may inspire future researches.
	\item Our experiments on CIFAR-10 and ImageNet have shown the effectiveness of AOFP in significantly reducing the parameters and FLOPs of several very deep off-the-shelf CNNs including ResNet-152.
	\item We propose Destructive CNN Re-design, a CNN design paradigm, which aims at optimizing the width of convolutional layers in order for higher accuracy and faster inference. E.g., the first two layers of VGG both have 64 filters, but we found out that 44 and 80 work better. This process can be used as a final refining step before a model is publicly released or deployed.
\end{itemize}

\section{Related Work}
Numerous researches \cite{lecun1990optimal,hassibi1993second,castellano1997iterative,han2015learning,guo2016dynamic} have proved it feasible to remove a large portion of parameters from neural networks without significant accuracy drop. Furthermore, by removing filters instead of sporadic connections from CNNs, we transform the wide convolutional layers into narrower ones to reduce the FLOPs, memory footprint and power consumption significantly.

A straightforward way of filter pruning is to remove unimportant filters from a well-trained model. Some researchers have discussed various metrics to measure filter importance. E.g., a prior work prunes filters by the accuracy reduction with the filter ablated \cite{abbasi2017structural}; magnitude-based pruning \cite{li2016pruning} considers the filters with larger magnitude more likely to be important; APoZ-based pruning \cite{hu2016network} calculates the percentage of zeros in the activated feature maps; some researchers use Taylor-expansions \cite{molchanov2016pruning}. There are also some inspiring works which pick up filters by no importance score but by the channel contribution variance \cite{polyak2015channel} or the Lasso regression \cite{he2017channel}.

A major drawback of the existing methods is the requirement for heuristic knowledge. \textbf{1)} The filter importance metrics \cite{li2016pruning,molchanov2016pruning,hu2016network} are heuristic, as it is not clear why the proposed metrics reflect the inherent importance of filters, and it is hard to judge if a heuristic method outperforms another. \textbf{2)} For iterative filter pruning methods, the granularity, i.e., the number of filters pruned at each step, is a key manually set hyper-parameter and a critical trade-off to be solved. The fewer filters are discarded once, the less damage is done to the model, which means the less finetuning time is required for the network to restore the accuracy; but more steps are needed to reach a satisfactory compression rate. \textbf{3)} It is difficult to decide when to stop pruning, i.e., the resulting width of each layer. Many works \cite{li2016pruning,hu2016network,he2017channel} have shown that some layers can be pruned by a large ratio without accuracy drop, but some layers are sensitive, which makes it hard to set layer-wise termination conditions.

Apart from pruning by importance, some other methods train the model under certain constraints (e.g., group Lasso \cite{roth2008group}) in order to zero out some filters \cite{alvarez2016learning,wen2016learning,ding2018auto} or make them identical for removal \cite{ding2019centripetal}. 

Moreover, some other CNN compression and acceleration techniques have also been intensively studied, including tensor low rank expansion \cite{jaderberg2014speeding}, parameter quantization \cite{han2015deep}, knowledge distillation \cite{hinton2015distilling}, DCT-based fast convolution \cite{wang2016cnnpack}, feature map compacting \cite{wang2017beyond}, etc. Of note is that AOFP is complementary to these methods.

\section{Approximated Oracle Filter Pruning}
\subsection{Formulation}
Let $i$ be the layer index, $\bm{M}^{(i)}\in\mathbb{R}^{h_{i}\times w_{i}\times c_{i}}$ be an $h_i\times w_i$ feature map with $c_{i}$ channels and $\bm{M}^{(i,j)}=\bm{M}^{(i)}_{:,:,j}$ be the $j$-th channel. The parameters of conv layer $i$ with kernel size $r_i\times s_i$ reside in the kernel tensor $\bm{K}^{(i)}\in \mathbb{R}^{r_{i}\times s_{i}\times c_{i-1}\times c_{i}}$ and the bias term $\bm{b}^{(i)}\in\mathbb{R}^{c_{i}}$, so we use $\bm{P}^{(i)}=(\bm{K}^{(i)},\bm{b}^{(i)})$ to denote the parameters of layer $i$. In this paper, filter $j$ at layer $i$ refers to the tuple comprising the trained parameters related to the output channel $j$ of layer $i$, $\bm{F}^{(i,j)}=(\bm{K}^{(i)}_{:,:,:,j},b^{(i)}_j)$, and we denote the set of all such filters at layer $i$ by $\mathcal{F}_i$. This layer takes $\bm{M}^{(i-1)}\in\mathbb{R}^{h_{i-1}\times w_{i-1}\times c_{i-1}}$ as input and outputs $\bm{M}^{(i)}$. Let $\ast$ be the 2-D convolution operator, an arbitrary output channel $j$ is
\begin{equation}
\bm{M}^{(i,j)}=\sigma^{(i)}((\sum_{k=1}^{c_{i-1}}\bm{M}^{(i-1,k)}\ast\bm{K}^{(i)}_{:,:,k,j})+b^{(i)}_j) \,,
\end{equation}
where $\bm{K}^{(i)}_{:,:,k,j}$ is the $k$-th input channel of the $j$-th filter, i.e., a 2-D convolution kernel, function $\sigma^{(i)}$ denotes the possible following operations such as non-linearities. For simplicity, we rewrite this transformation as a function $\zeta^{(i)}$,
\begin{equation}\label{eq-conv-transformation}
\bm{M}^{(i)}=\zeta^{(i)}(\bm{M}^{(i-1)},\mathcal{F}_i) \,.
\end{equation}

Importance-based filter pruning methods define the importance of filters in terms of some measures, score the filters by some means, then prune the unimportant parts and reconstruct the network using the remaining parameters. Let $T$ be the filter importance score value, $\delta$ be the threshold and $\mathcal{I}_i$ be the filter index set of layer $i$ (e.g., if conv5 has four filters, then $\mathcal{I}_5=\{1,2,3,4\}$), the remaining set, i.e., the index set of the filters which survive the pruning, is $\mathcal{R}_i=\{j\in \mathcal{I}_i \ |\ T(\bm{F}^{(i,j)})>\delta\}$. We prune the other filters by reconstructing the network using the remaining parameters sliced from the original kernel and bias term. That is,
\begin{equation}\label{eq4}
\bm{P}^{(i)} \gets (\bm{K}^{(i)}_{:,:,:,\mathcal{R}_i},\bm{b}^{(i)}_{\mathcal{R}_i}) \,.
\end{equation}
If the conv layer is followed by a batch normalization \cite{ioffe2015batch} layer, its parameters should be handled in the same way as the bias term $\bm{b}$. The input channels of the following layer corresponding to the pruned filters should also be discarded,
\begin{equation}\label{eq6}
\bm{P}^{(i+1)} \gets (\bm{K}^{(i+1)}_{:,:,\mathcal{R}_i,:},\bm{b}^{(i+1)}) \,.
\end{equation}

\subsection{Rethinking Oracle Pruning}
In this subsection, we focus on the situation where we prune $q$ filters from a layer which originally has $c$ filters. Oracle Pruning assesses a filter's importance by looking at the model's accuracy drop when the filter is ablated. Formally, let $\mathcal{F}$ be the original filter set of the CNN, $L(x,y,\mathcal{F})$ be the accuracy-related loss value (e.g., cross-entropy loss for classification tasks) generated with the given filter set, $X$ and $Y$ be the examples and labels of the assessment dataset, which is a subset of the training dataset, the \textit{scoring process} aims to obtain the importance score for each filter $\bm{F}$ by
\begin{equation}
T(\bm{F}) = \sum_{(x,y) \in (X,Y)} (L(x,y,\mathcal{F}-\bm{F}) - L(x,y,\mathcal{F})) \,,
\end{equation}
where $L(x,y,\mathcal{F}-\bm{F})$ is the loss value computed without filter $\bm{F}$, i.e., with the corresponding feature map channel removed or equivalently masked out. In this way, we can slim a layer by always removing the filter with the least $T$ value and re-scoring the remaining filters for $q$ times. Compared to the heuristic approaches, where $T$ is computed in other ways, an obvious strength of Oracle Pruning is the accuracy, while its weakness is the high time complexity. Specifically, using an assessment dataset of $\gamma$ examples, the time complexity of Oracle Pruning is $O(cq\gamma)$, because we need to ablate every remaining filter in turn ($O(c)$) then test on the assessment dataset ($O(\gamma)$) to pick up a single filter to prune, and this scoring process is conducted for $q$ times.

A straightforward way to alleviate the computational burdens is to prune several filters once at a time, trading accuracy for efficiency. However, as all the filters at a conv layer compose a highly non-linear system \cite{mozer1989using}, removing a filter can affect the relative importance of other filters, inevitably resulting in poor accuracy (Fig. \ref{fig-compre-pruning}). We refer to the number of filters pruned at a time as the \textit{granularity} $g$. Except for lower accuracy, another downside of granular pruning is that we introduce an extra hyper-parameter $g$, which may require heuristic knowledge and human efforts to tune. For example, we can estimate the redundancy of a layer by pruning some filters and observing the accuracy drop in advance, such that we set $g$ to a larger value to reduce the time cost if the layer seems to be highly redundant, or adopt a smaller $g$ to prune more carefully.

\subsection{Damage Isolation}
The essence of Oracle Pruning is to observe the consequences of the temporary removal of filters, i.e., to observe the feedback of many pruning attempts, which is generated by computing the final loss value. In this way, even when we are trying to prune a low-level layer, we still need to feed the input data through the entire network to obtain the feedback. Even worse, using such a feedback loop, we can only deal with one layer at a time, because as we ablate the filters in a specific layer, the subsequent information flow of the network is changed, such that the scoring processes of the higher-level layers are affected. Therefore, we seek to shorten the feedback loop for faster inference and mutually independent parallel filter scoring on every layer. 

Our proposed approach is based on an intuition that a CNN can be viewed as a state machine, where the feature maps (states) are transformed by the operations performed by conv layers (Eq. \ref{eq-conv-transformation}). So essentially, the change in the filters at layer $i$, i.e., the change in $\bm{M}^{(i)}$, is isolated by the subsequent layer $i+1$, because layer $i+2$ and the higher-level layers cannot see the change in $\bm{M}^{(i)}$. Taking the extreme case for example, if we prune some filters at layer $i$, but observe no difference in $\bm{M}^{(i+1)}$, we can claim that the pruning does no damage to the model because the input states to the remaining part of the network are not changed. Inspired by this, we propose to calculate the approximated importance score $T^\prime$ based on the output of the next layer,
\begin{equation}\label{approximated-T-prime}
T^\prime(\bm{F}) = \frac{1}{|X|} \sum_{x \in X} t(\bm{F}, x) \,,
\end{equation}
where $\bm{F}$ is a filter at layer $i$, $t$ is the \textit{isolated damage sample} which reflects how much the output of layer $i+1$ on input example $x$ is deviated by the pruning attempt on $\bm{F}$,
\begin{equation}\label{eq-compute-t}
t(\bm{F}, x) = \frac{||\bm{M}^{(i+1)}(x) - \zeta^{(i+1)}(\bm{M}_{\bm{F}}^{(i)}(x), \mathcal{F}^{(i+1)})||_{2}^{2}}{||\bm{M}^{(i+1)}(x)||_2^2} \,.
\end{equation}
Here $\bm{M}_{\bm{F}}^{(i)}(x)$ is the output of layer $i$ derived without $\bm{F}$,
\begin{equation}
\bm{M}_{\bm{F}}^{(i)}(x)=\zeta^{(i)}(\bm{M}^{(i-1)}(x),\mathcal{F}_i-\bm{F}) \,.
\end{equation}
Except for Euclidean distance, other distance functions may work as well, which are beyond the scope of this paper.

\subsection{Multi-path Training-time Pruning Framework}
It is common to finetune the whole model after each time of pruning \cite{li2016pruning,molchanov2016pruning,hu2016network,abbasi2017structural}, i.e., the scoring and finetuning processes are serial. To reduce the time cost, we propose a multi-path training-time pruning framework (Fig. \ref{framework}) to parallelize the scoring and finetuning.

Specifically, when we prune a certain conv layer $i$, the computation flow after it is split into two paths, which are referred to as the base path and the scoring path, respectively. E.g., Fig. \ref{framework} shows two scoring paths which each contain only one conv layer (conv2, conv3) as we are pruning conv1 and conv2 simultaneously. The base path forwards the outputs of layer $i$ through a \textit{base mask} $\bm{u}^{(i)}\in \mathbb{R}^{c_i}$ initialized as $\bm{1}$. The $j$-th channel of the output of the next layer becomes
\begin{equation}
\bm{M}^{(i+1,j)}=\sigma^{(i+1)}((\sum_{k=1}^{c_{i}}u^{(i)}_k \bm{M}^{(i,k)}\ast\bm{K}^{(i+1)}_{:,:,k,j})+b^{(i+1)}_j) \,.
\end{equation}
It is obvious that setting $u^{(i)}_k=0$ is equivalent to removing the $k$-th filter at layer $i$. At the endpoint of the base path, the original loss value is calculated, the gradients are derived and the model parameters are updated. Meanwhile, the scoring path goes through a \textit{scoring mask} $\bm{v}^{(i)} \in \mathbb{R}^{c_i}$, the masked $\bm{M}^{(i)}$ is fed into layer $i+1$, then the isolated damage sample $t$ is computed and stored in memory. 

During the training process, for each batch of input data, we randomly set some bits in the scoring mask to zero, such that the corresponding filters are ablated on the scoring path but still kept on the base path. The $t$ value is computed by comparing the corresponding feature maps on the base and scoring paths, and if it is large, we learn that the ablated filters are important for the current batch of data. With more and more samples collected, we become more and more confident to tell which filters are the least important. When enough samples have been collected, we approximate $T^\prime$ for each filter $j$ in the layer by
\begin{equation}\label{approximated-T-hat}
\hat{T}(\bm{F}^{(i,j)})=mean(\mathcal{T} ^{(i,j)}) \,,
\end{equation}
where $\mathcal{T} ^{(i,j)}$ is a set which records all the $t$ values collected with filter $j$ ablated.

With all the filters scored, we pick up $g$ filters with the lowest $\hat{T}$ value, fix the corresponding bits in $\bm{u}^{(i)}$ and $\bm{v}^{(i)}$ to zero, such that the filters are masked out permanently. Of note is that changing some bits in $\bm{u}^{(i)}$ affects the back-propagation through the base path, thus the network's accuracy will be restored by finetuning. In the meantime as finetuning, we choose the next $g$ filters to prune through the scoring path. We refer to the process of choosing one or more filters to prune (in the meantime as recovering the damage caused by the last pruning) as a \textit{move}. When some termination conditions have been met, we remove the filters according to the base mask by Eq. \ref{eq4}, \ref{eq6} with $\mathcal{R}_i=\{j|u^{(i)}_j=1\}$, such that the layer is slimmed with no further accuracy drop.

Such a multi-path framework enables parallel scoring and pruning on multiple layers, i.e., we can prune layer $i$ according to the outputs of layer $i+1$ and prune layer $i+1$ according to layer $i+2$ simultaneously. Scoring a low-level conv layer does not affect the higher-level ones because every scoring process compares the outputs of the scoring path with the base path unaffected by the pruning attempts (i.e., the changes of the scoring masks) on the previous layers.

Note that though we randomly mask out channels in a dropout-like \cite{srivastava2014dropout,molchanov2017variational} manner, we do not rescale the remaining parts for compensation as we do when using dropout for regularization.

\subsection{Binary Filter Search}
\begin{algorithm}[t]
	\caption{Approximated Oracle Filter Pruning}
	\label{alg}
	\begin{algorithmic}[1]
		\STATE {\bfseries Input:} the target layer $i$ of the original CNN, refinement threshold $\theta$, search cost $\phi$
		\STATE Base mask $\bm{u}\gets\bm{1}$
		\WHILE {True}
		\STATE Search space $\mathcal{A}\gets\{j | u_j=1\}$
		\REPEAT
		\STATE Loss record set $\mathcal{T}^{(i,j)}\gets\{\} ,\,\forall\, j\in\mathcal{A}$
		\REPEAT
		\STATE Randomly choose $|\mathcal{A}|/2$ elements out of $\mathcal{A}$ as the ablated filter index set $\mathcal{H}$
		\STATE Initialize $\bm{v}\gets\bm{u}$, let $v_j\gets0, \,\forall\, j \in \mathcal{H}$
		\STATE Generate and forward a batch of input data, compute the $t$ value as Eq. \ref{eq-compute-t}, record it for the ablated filters by
		$\mathcal{T}^{(i,j)}\gets\mathcal{T}^{(i,j)}\cup\{t\} \,,\,\forall\, j \in \mathcal{H}$
		\STATE Back-prop gradients, update parameters
		\UNTIL {$\phi$} batches have been forwarded
		\STATE Compute $\hat{T}$ for each filter $j \in \mathcal{A}$ as Eq. \ref{approximated-T-hat}
		\STATE Pick up $|\mathcal{A}|/2$ filters with the smallest $\hat{T}$ value as the picked set $\mathcal{B}$
		\STATE Max damage $p=max(\{\hat{T}(\bm{F}^{(i,j)}) \,|\, \forall\, j \in \mathcal{B}\})$
		\STATE Let $\mathcal{A}\gets\mathcal{B}$
		\UNTIL $p < \theta$ or $|\mathcal{B}| = 1$
		\IF { $p < \theta \,,$}
		\STATE Let $\bm{u}_j\gets0, \,\forall\, j \in \mathcal{B}$ \quad // prune the picked filters
		\ELSE
		\STATE {\bfseries break} \quad // $p\geq\theta$ and $|\mathcal{B}|=1$, stop refining
		\ENDIF
		\ENDWHILE
		\STATE Prune layer $i$ by Eq. \ref{eq4}, \ref{eq6} with $\mathcal{R}_i=\{j|u_j=1\}$
		\STATE {\bfseries Return}
	\end{algorithmic}
\end{algorithm}
In this subsection, we discuss and solve three problems of the proposed framework. \textbf{1)} On a specific layer, the finetuning process makes the parallel scoring inaccurate. When $g$ filters have been masked out permanently, i.e., the corresponding bits in the two masks have been fixed to zero, the network needs a period to recover, during which the filter importance assessment is not accurate. Namely, since the remaining filters vary during the finetuning period after the last pruning, the $t$ values obtained in this period do not accurately reflect the actual importance of the filters which are being scored. \textbf{2)} The optimal value of the granularity $g$ is hard to resolve. \textbf{3)} It is heuristic to decide when to stop pruning, as we do not know the optimal resulting width.

Inspired by the idea of \textit{incremental refinement}, we propose to search for the least important filters in a binary search manner. Concretely, at the beginning of each move, all the remaining filters compose the search space $\mathcal{A}$. We first score every filter in $\mathcal{A}$ and pick up $|\mathcal{A}|/2$ filters as the picked set $\mathcal{B}$ which are most likely to be unimportant. Though the network has not become stable (if this is not the first move), the assessment is not accurate indeed, but accurate enough for such a coarse-grained search. When $\mathcal{B}$ is obtained, the network finetuned through the base path has become more stable, so we abandon the collected samples and start a more fine-grained search by letting $\mathcal{A} \gets \mathcal{B}$, searching for the less important half of the picked set (i.e., a quarter of the last search space). As we use imprecise samples to do coarse searches and high-quality samples to search accurately, the samples collected in the meantime as finetuning are fully utilized, and the accuracy of importance scoring is guaranteed. We refer to the number of collected $t$ samples needed to complete one step of binary search as the \textit{search cost} $\phi$.
\begin{figure}
	\begin{center}
		\includegraphics[width=0.9\columnwidth]{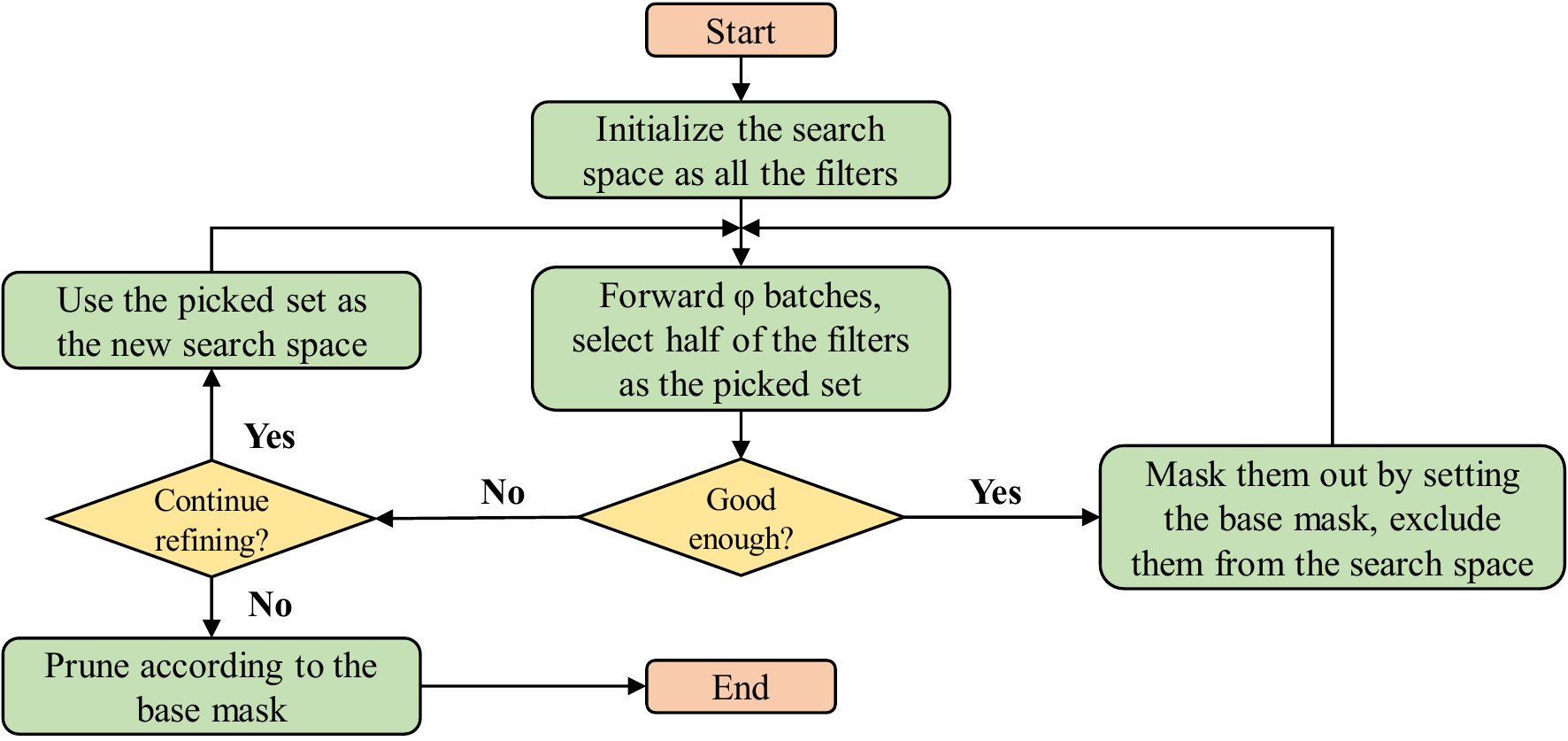}
		\caption{Flow chart of AOFP on a single layer.}
		\label{flow-chart}
	\end{center}
	\vskip -0.1in
\end{figure}

Except for accurate assessment, Binary Filter Search also helps the decision of the granularity $g$ and the judgment of the termination conditions in a natural way, freeing us from heavy works on layer sensitivity analysis experiments \cite{he2017channel,li2016pruning,hu2016network} and heuristic manually set controlling conditions. Essentially, at each step of binary search, the picked set can be regarded as the least important $|\mathcal{B}|$ filters. If we finish the current move by pruning them, $|\mathcal{B}|$ serves exactly as the granularity $g$. So in the context of binary search, the problem of deciding $g$ and the termination conditions can be simply solved as follows: 
\begin{itemize}[noitemsep,nolistsep,,topsep=0pt,parsep=0pt,partopsep=0pt]
	\item If the current picked set $\mathcal{B}$ is good enough, finish the current move with $g=|\mathcal{B}|$ (i.e., permanently mask out the filters in $\mathcal{B}$ and start a new move); otherwise, see if it is possible to continue refining ($|\mathcal{B}|>1$ or $|\mathcal{B}|=1$).
	\item If $|\mathcal{B}|>1$, we continue refining by letting $\mathcal{A} \gets \mathcal{B}$; otherwise, it suggests that the single least important filter is still too important, so we stop pruning the layer.
\end{itemize}

We introduce a global hyper-parameter, the \textit{refinement threshold} $\theta$, which is used to compare with the max $\hat{T}$ value (Eq. \ref{approximated-T-hat}) of the filters in $\mathcal{B}$ to judge if the picked set is good (unimportant) enough. We say $\mathcal{B}$ is good enough if
\begin{equation}
max(\{\hat{T}(\bm{F}^{(i,j)}) \,|\, \forall\, j \in \mathcal{B}\}) < \theta \,.
\end{equation}
Intuitively, $\theta$ indicates the upper limit of the damage we can endure for a single step of pruning. E.g., with $\theta=0.02$, we consider it acceptable to prune one or more filters with 2\% isolated damage. With a larger $\theta$, we tend to prune with larger granularity, and vice versa.

In this way, another design concern is settled naturally, that is, how many filters to randomly ablate for a batch of input data. We randomly ablate $|\mathcal{A}|/2$ filters out of the search space $\mathcal{A}$ at a time, and the reason is simple: according to our discussions above, the collected $t$ values should reflect the expected damage if we prune the current picked set, i.e., the $t$ values should be derived with $|\mathcal{B}|$ filters ablated, and $|\mathcal{B}|=|\mathcal{A}|/2$. The AOFP algorithm on a single layer is outlined in Fig. \ref{flow-chart} and formally described in Alg. \ref{alg}. In practice, we apply AOFP on every layer simultaneously.

\section{Experiments}
\subsection{Comparison of Filter Pruning Metrics}\label{sec-exp-1}
\begin{figure}
	\begin{center}
		\includegraphics[width=0.8\linewidth]{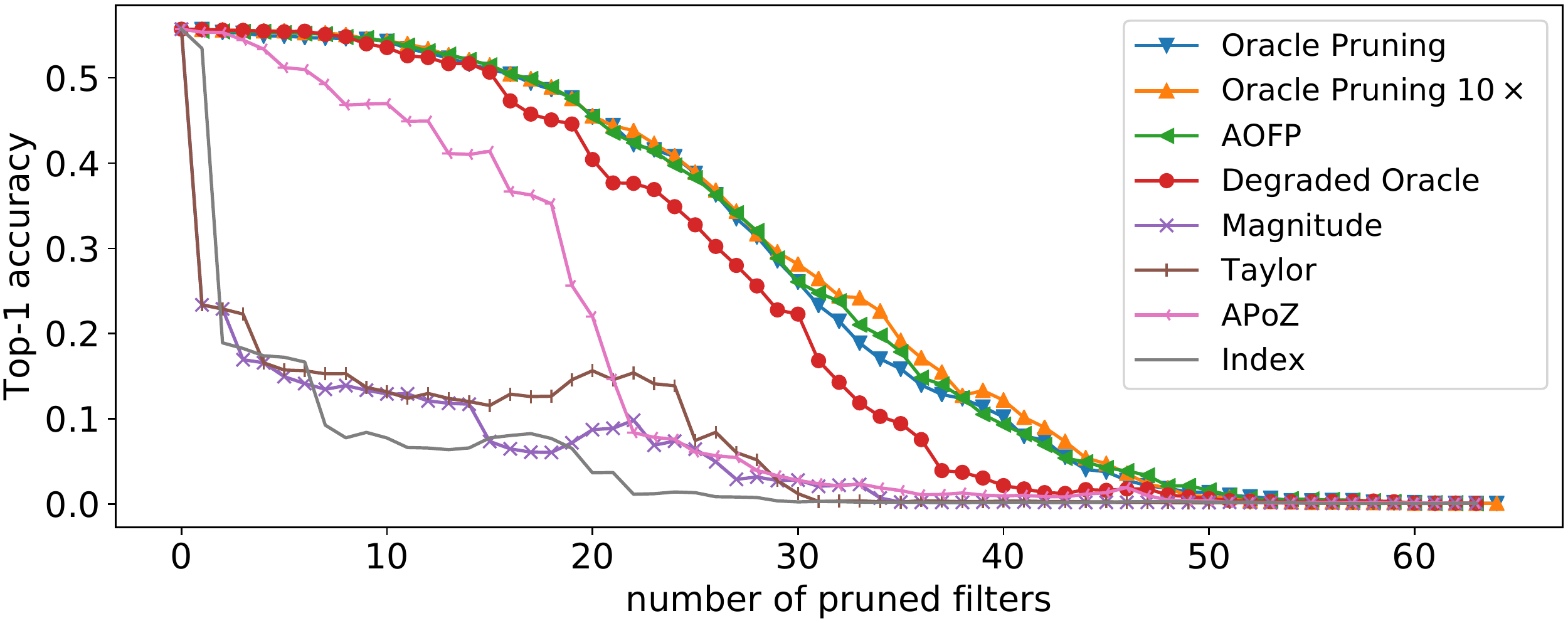} 
		\caption{Comparison of filter pruning metrics on AlexNet.}
		\label{fig-compre-pruning}
	\end{center}
	\vskip -0.1in
\end{figure}
In this subsection, we present a comparison of Oracle Pruning, AOFP and other heuristic methods using AlexNet \cite{krizhevsky2012imagenet} on ImageNet \cite{deng2009imagenet}. We use the simplified version of AlexNet \cite{Tensorflow-AlexNet}, which is composed of five stacked conv layers and three fully-connected layers with no LRU or cross-GPU connections. For faster convergence, batch normalization \cite{ioffe2015batch} is applied on every conv layer. We compare these metrics by pruning filters one by one from the first layer without finetuning. Fig. \ref{fig-compre-pruning} shows the Top-1 accuracy on the validation set with varying number of pruned filters. For Oracle Pruning, APoZ-based and Taylor-expansion-based, we use randomly sampled 10,000 training examples as the assessment dataset. For AOFP, we set the search cost $\phi$ such that the total number of examples consumed equals that of Oracle Pruning. As we are pruning only one layer, we apply Binary Filter Search to collect the final loss value instead of the isolated damage. For Oracle Pruning $10\times$, we use 100,000 examples to score a filter. For Degraded Oracle, no re-scoring processes are conducted, i.e., the importance scores collected at the very beginning are used to guide the pruning till the end. For the curve labeled as Index, we prune filters from the first one to the 64th, which is essentially equivalent to random guess. 

Our observations are as follows. \textbf{1)} By comparing Degraded Oracle and Oracle Pruning, we conclude that re-computing importance scores after each step of Oracle Pruning is essential, as a CNN is a highly non-linear system, and the removal of a filter can affect the relative importance of other filters. This discovery is consistent with the observations of prior works \cite{mozer1989using,sharma2017incredible} that neural networks do not distribute the learning representation equitably across neurons. \textbf{2)} AOFP is almost as good as Oracle Pruning. \textbf{3)} The extra costs of Oracle Pruning $10\times$ bring marginal accuracy improvement.

\subsection{AOFP for Automatic CNN Compression}\label{sec-exp-2}
Abundant experiments are conducted using several common networks on CIFAR-10 \cite{krizhevsky2009learning} and ImageNet, including AlexNet, VGG, and ResNets, to validate the effectiveness of AOFP, which is measured by the FLOPs and accuracy of the pruned model. For reproducibility and comparability, we use the same VGG version as other works \cite{li2016pruning, liu2017learning}, and the same ResNet structures as the official tensorflow/slim models \cite{Tensorflow-slim}. All the original models are trained from scratch, and the FLOPs of every model are calculated in the same manner as \cite{he2016deep}. Of note is that we perform AOFP on every target layer simultaneously. Concretely, for a specific layer, when the returning condition in Alg. \ref{alg} is satisfied, the AOFP process is restarted, i.e., we finish the current move without changing the base mask and start the next move. After each move on any conv layer, we calculate the reduced global FLOPs based on the model architecture and the current values of all the base masks, and stop pruning when the reduced FLOPs reach a target level (e.g., above 60\% for the model labeled as AOFP-A3 in Table. \ref{exp-table-cifar10}). Then we reconstruct a narrower network following Eq. \ref{eq4}, \ref{eq6} for every layer, finetune the model, and test it on the validation dataset using a single central crop.

On CIFAR-10, we use VGG-16 for a quick sanity check. The base model is trained from scratch for 600 epochs to ensure the convergence, with the standard data augmentation, i.e., padding to $40\times40$, random cropping and flipping. We use a batch size of 64 and a learning rate initialized to $5\times10^{-2}$ then decayed by 0.1 every 200 epochs. We perform AOFP on all of the 13 layers with search cost $\phi=20,000$, $\theta=0.01$ and a constant learning rate of $1\times10^{-3}$. On ImageNet, we first prune all the conv layers of AlexNet with $\phi=4,000$, $\theta=0.02$ and a learning rate of $1\times10^{-3}$. For ResNets, since there are more layers being simultaneously pruned, we increase the search cost to $\phi=8,000$ for better filter scoring and damage recovery. These hyper-parameters are casually set without careful tuning. Of note is that, on ResNets, due to the constraints of the shortcut connections, only the internal layers (i.e., the first and second layers in each block which are not directly added to the identity mapping) are pruned, as a common practice \cite{luo2017thinet,luo2018autopruner,wang2017structured}.
\begin{table}[t]
	\caption{Pruned VGG on CIFAR-10. The resulting models with different FLOPs are labeled from A1 to A5.}
	\label{exp-table-cifar10}
	\begin{center}
		\begin{small}
			\begin{tabular}{llccccccc}
				\toprule
				Result 	& Top-1 	& FLOPs & FLOPs$\downarrow$\%	\\
				\midrule
				base	&	93.38	&	313M&	-		\\
				\textbf{AOFP-A1}	&	\textbf{93.81}	&	\textbf{215M}&	\textbf{31.32}	\\
				\cite{li2016pruning}	&	93.40	&	206M	&34.20	\\
				\textbf{AOFP-A2}	&	\textbf{94.03}	&	\textbf{186M}&	\textbf{40.51}	\\
				\cite{liu2017learning}			&	93.80	&	-		&51.00	\\
				\cite{huang2018learning} &	91.67	&	-		&55.20	\\
				\textbf{AOFP-A3}	&	\textbf{93.84}	&	\textbf{124M}&	\textbf{60.17}	\\
				\cite{xu2018globally}			&	93.29	&	120M	&61.46	\\	
				\textbf{AOFP-A4}	&	\textbf{93.47}	&	\textbf{108M}&	\textbf{65.27}	\\
				\cite{zhou2018online}			&	92.33	&	-		&74.81	\\
				\textbf{AOFP-A5}	&	\textbf{93.28}	&	\textbf{77M}&	\textbf{75.27}	\\
				\bottomrule
			\end{tabular}
		\end{small}
	\end{center}
\vskip -0.1in
\end{table}
\begin{table}[t]
	\caption{Pruning on ImageNet. The competitors include ThiNet \cite{luo2017thinet}, NISP \cite{yu2018nisp}, Channel Pruning \cite{he2017channel}, SPP \cite{wang2017structured}, Autopruner \cite{luo2018autopruner}, ISTA-based \cite{ye2018rethinking}, C-SGD \cite{ding2019centripetal}.}
	\label{exp-table-imagenet}
	\begin{center}
		\begin{small}
			\begin{tabular}{llccccccc}
				\toprule
				& Result 	& Top-1 & Top-5	& FLOPs & $\downarrow$\%	\\
				\midrule
				Alex	&	base	&	55.71	& 79.45 &	838M	&	-		\\
				\textbf{Alex}	&	\textbf{AOFP-B1}	&	\textbf{56.54}	& \textbf{79.95} &	\textbf{578M}	&	\textbf{30.98}	\\
				\textbf{Alex}	&	\textbf{AOFP-B2}	&	\textbf{56.17}	& \textbf{79.53} &	\textbf{492M}	&	\textbf{41.33}	\\
				\midrule
				Res50	&	base	&	75.34	& 92.56 &	3.85G	&	-		\\
				\textbf{Res50}	&	\textbf{AOFP-C1}	&	\textbf{75.63}	& \textbf{92.69} &	\textbf{2.58G}	&	\textbf{32.88}	\\
				Res50	&	ThiNet-70&	72.04	& 90.67	&	2.44G	&	36.75	\\
				Res50	&	NISP	&	0.89$\downarrow$	&-	&-	&	44.01	\\
				Res50	&	Chan-Pr	&-		&90.80	&	-		&	50.00	\\
				Res50	&	SPP			&-		&90.40	&	-		&	50.00	\\
				Res50	&	Autopr		&74.76	&92.15	&	-		&	51.21	\\
				Res50	&	ThiNet-50&	71.01	& 90.02	&	1.70G	&	55.76	\\
				Res50	&	C-SGD-50	&	74.54	& 92.09	&	1.70G	&	55.76	\\
				\textbf{Res50}	&	\textbf{AOFP-C2}	&	\textbf{75.11}	& \textbf{92.28} &	\textbf{1.66G}	&	\textbf{56.73}	\\
				\midrule
				Res101	&	base	&	76.63	& 93.29 &	7.57G	&	-		\\
				\textbf{Res101}	&	\textbf{AOFP-D1}	&	\textbf{76.88}	& \textbf{93.49} &	\textbf{5.29G}	&	\textbf{30.11}	\\
				Res101	&	ISTA	&	75.27	& -		&	4.47G	&	40.95	\\
				\textbf{Res101}	&	\textbf{AOFP-D2}	&	\textbf{76.40}	& \textbf{93.07} &	\textbf{3.77G}	&	\textbf{50.19}	\\

				\midrule
				Res152	&	base	&	77.37	& 93.52	&	11.28G	&	-		\\
				\textbf{Res152}	&	\textbf{AOFP-E1}	&	\textbf{77.47}	& \textbf{93.76} &	\textbf{6.12G}	&	\textbf{45.72}	\\
				\textbf{Res152}	&	\textbf{AOFP-E2}	&	\textbf{77.00}	& \textbf{93.49} &	\textbf{4.13G}	&	\textbf{63.36}	\\
				\textbf{Res152}	&	\textbf{AOFP-E3}	&	\textbf{76.40}	& \textbf{93.02} &	\textbf{2.85G}	&	\textbf{74.69}	\\
				\bottomrule
			\end{tabular}
		\end{small}
	\end{center}
\end{table}

As it turns out (Table. \ref{exp-table-cifar10}, \ref{exp-table-imagenet}), these networks can be pruned significantly even with an increase in accuracy due to the optimized network structure, which is consistent with the observations in other works \cite{liu2017learning,li2016pruning}. And if we wish to trade accuracy for efficiency, AOFP can reduce the computational burdens by a large margin, demonstrating not only higher pruning ratios but also less accuracy drop than other methods. Moreover, the increased network depth does not hinder the application of AOFP, because we simultaneously prune every layer in a mutually interdependent manner, and do not suffer from the notorious problem of error propagation and amplification in filter importance estimation \cite{yu2018nisp}, thanks to Damage Isolation.

\begin{figure}[t] 
	\begin{subfigure}{0.49\columnwidth}\label{width-plot-vc}
		\includegraphics[width=1\linewidth]{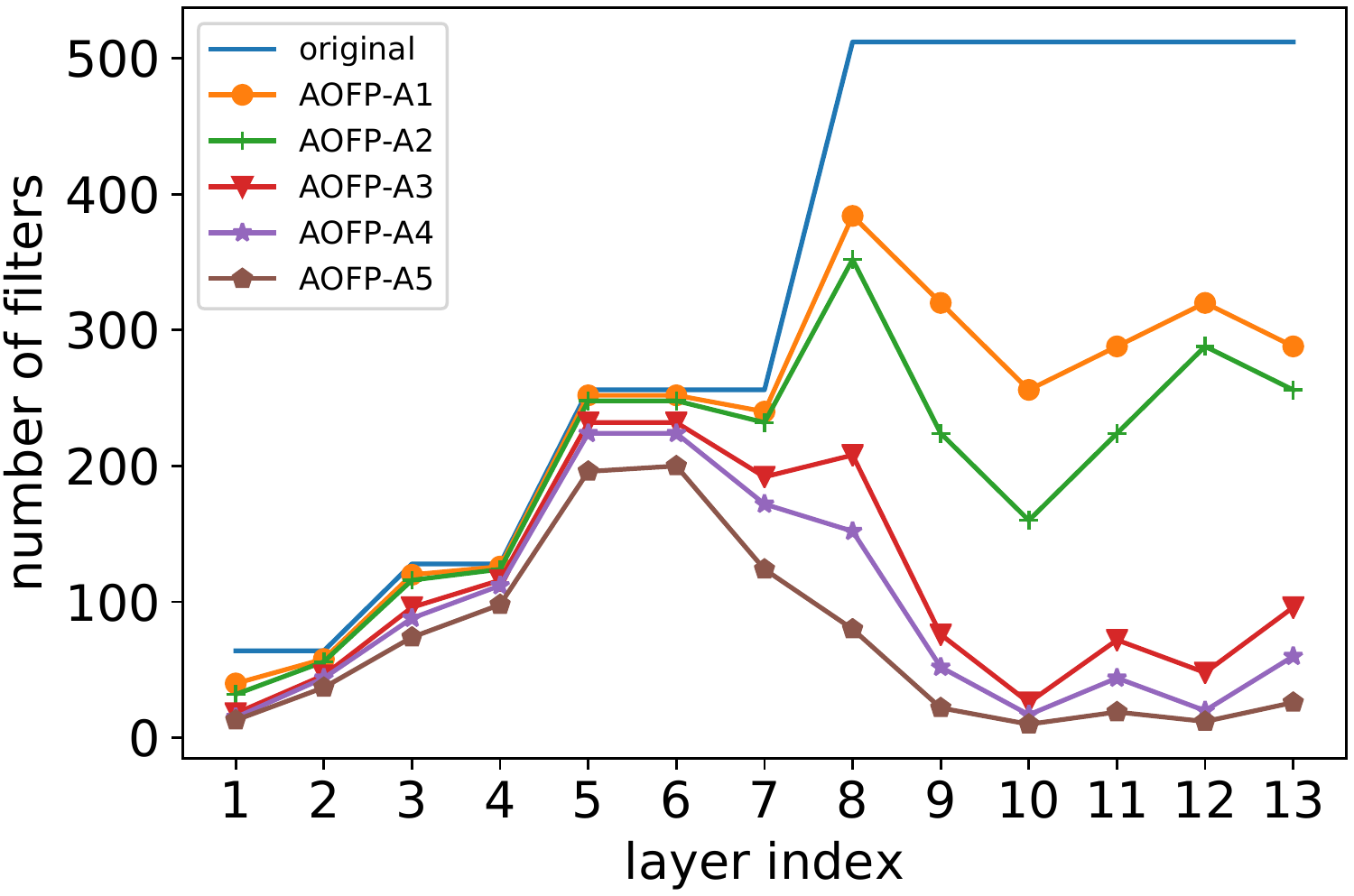} 
	\end{subfigure}
	\begin{subfigure}{0.49\columnwidth}\label{width-plot-resnet152}
		\includegraphics[width=1\linewidth]{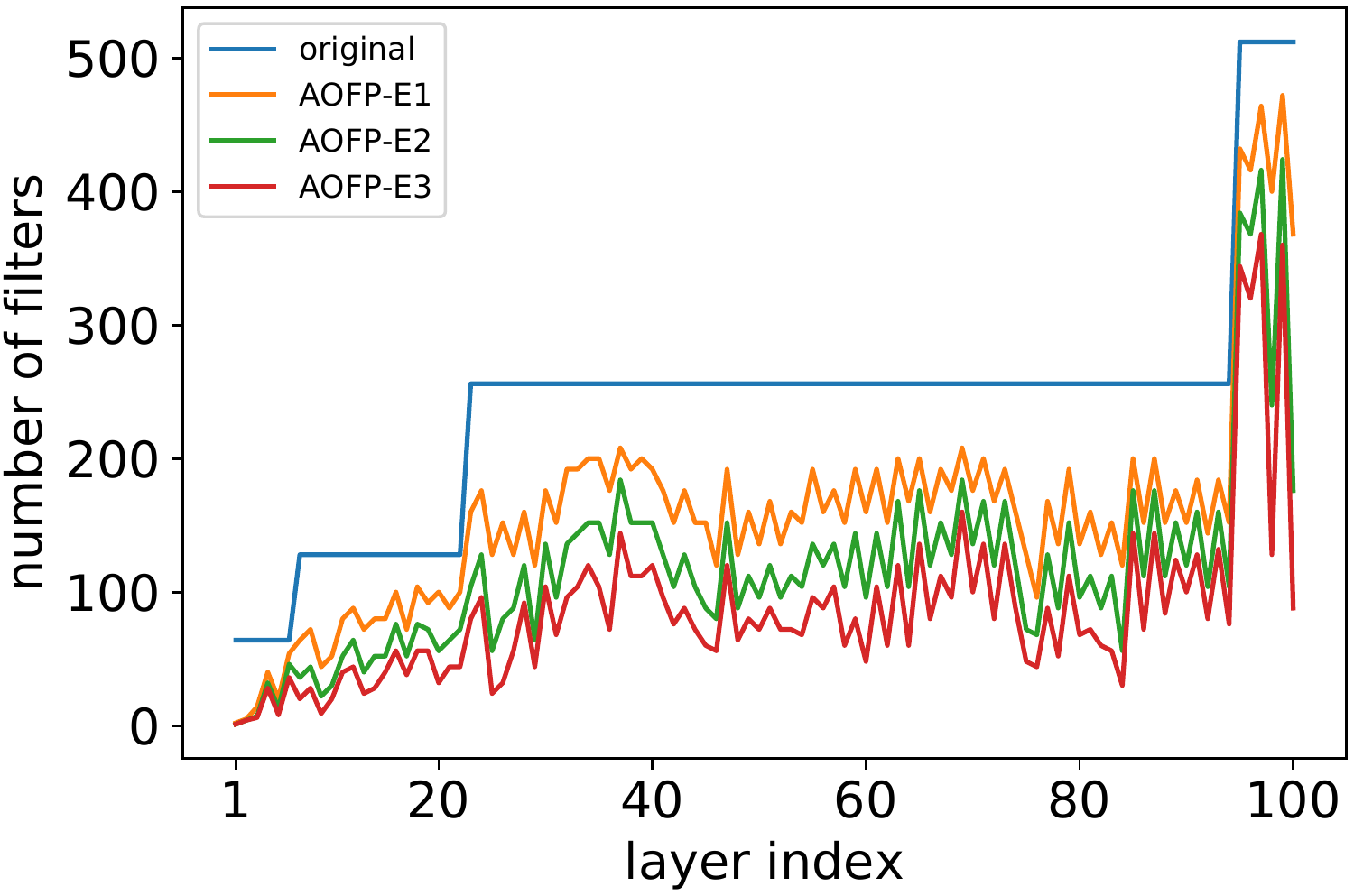} 
	\end{subfigure}
	\caption{Layer width of pruned models. Left: VGG on CIFAR-10. Right: ResNet-152 on ImageNet (only the pruned layers).}
	\label{width-plot}
\end{figure}
\begin{table}[t]
	\caption{AOFP pruned v.s. uniformly slimmed VGG. All the models are tested on an Nvidia GTX 1080Ti GPU or E5-2680 CPU with batch size 64, measured in examples/sec.}
	\label{exp-table-uniformly-slimmed}
	\begin{center}
		\begin{small}
			\begin{tabular}{llccccccc}
				\toprule
				 					& Top-1 	& FLOPs		& CPU 	& GPU		&	Speedup	\\
								\midrule
				VGG base			&	93.38	&	313M	& 343	&	6336	&	-				\\
				\textbf{AOFP-A3}				&	\textbf{93.84}	&	\textbf{124M}	& \textbf{683}	&	\textbf{13903}	&	$\bm{2.19\times}$	\\
				Uniform				&	92.88	&	126M	& 560	&	10224	&	$1.61\times$	\\
				\bottomrule
			\end{tabular}
		\end{small}
	\end{center}
\end{table}
Of note is that AOFP automatically detects the easy-to-prune layers without heuristic knowledge or manually set control conditions, which is a significant strength compared to other approaches where we have to empirically decide the width of every layer in advance \cite{li2016pruning, luo2017thinet, wang2017structured}. By visualizing the structure of the pruned networks in Fig. \ref{width-plot}, we learn that conv2,4,5,6 of VGG are more sensitive to pruning, but conv1 and the top six layers can be pruned dramatically for free, as AOFP chooses to prune these layers aggressively to achieve high pruning ratios. Similarly, AOFP converts the original tidy ResNet-152 to a more efficient one without human intervention. One concern about the irregularly shaped models is that the varying width of layers may cause GPU memory bottlenecks, so it may not result in real acceleration, though the FLOPs are reduced. However, our pruned VGG outperforms a uniformly slimmed counterpart in both accuracy and speed (Table. \ref{exp-table-uniformly-slimmed}). Concretely, we construct a VGG model where every layer is 69\% of its original width, such that the FLOPs becomes 126M, which is comparable to our pruned model labeled as AOFP-A3. We train it from scratch for 600 epochs and test it on both CPU and GPU. It is not clear why AOFP-A3 runs faster than the counterpart, but evidently, the discrimination towards irregularly shaped CNNs is just a kind of stereotype.

\subsection{AOFP for Global Progressive Pruning}
\begin{figure}[t] 
	\begin{subfigure}{0.49\columnwidth}\label{fig-resnet152-four-layers}
		\includegraphics[width=1\linewidth]{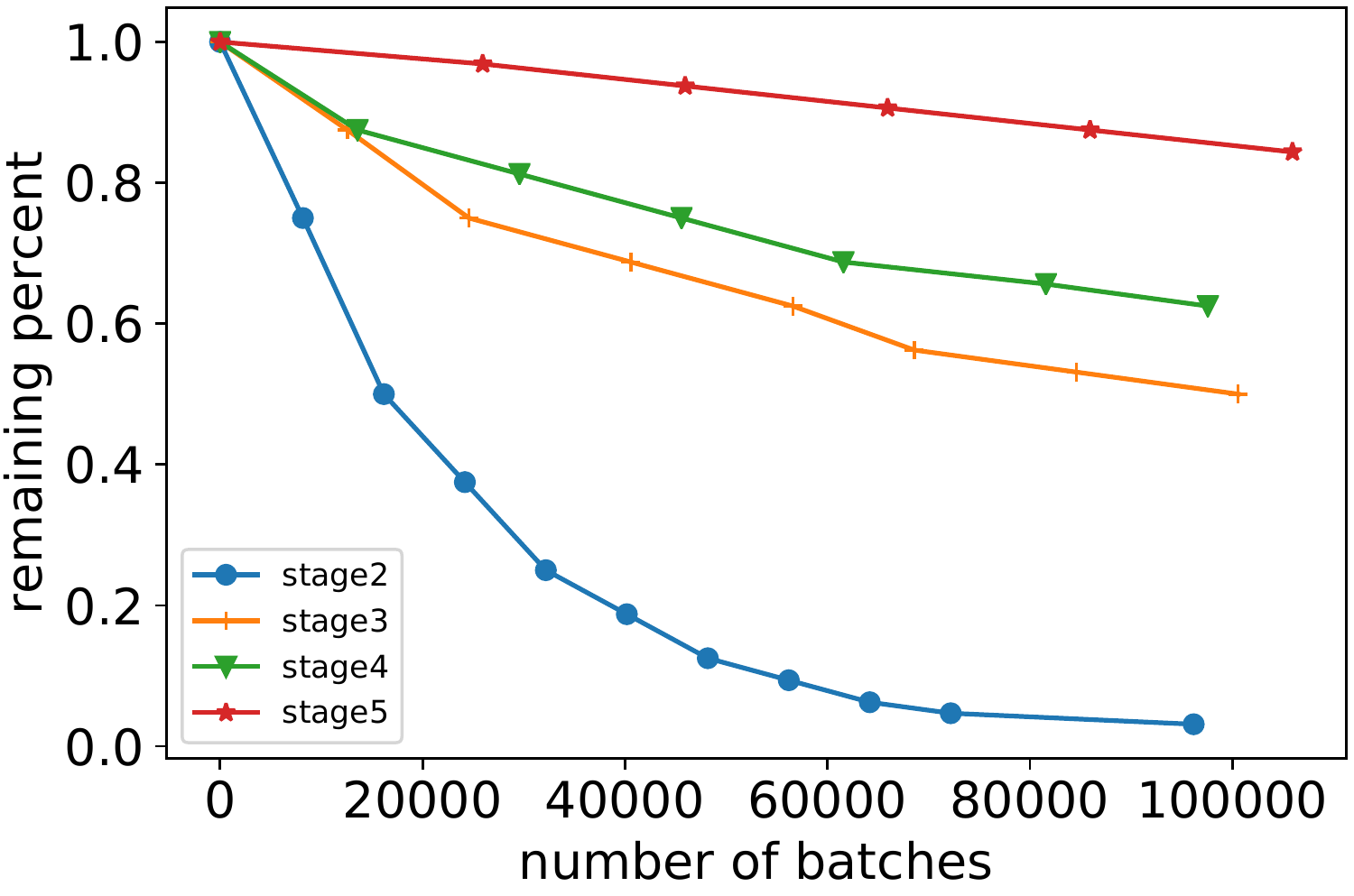} 
	\end{subfigure}
	\begin{subfigure}{0.49\columnwidth}\label{fig-resnet152-width-batches}
		\includegraphics[width=1\linewidth]{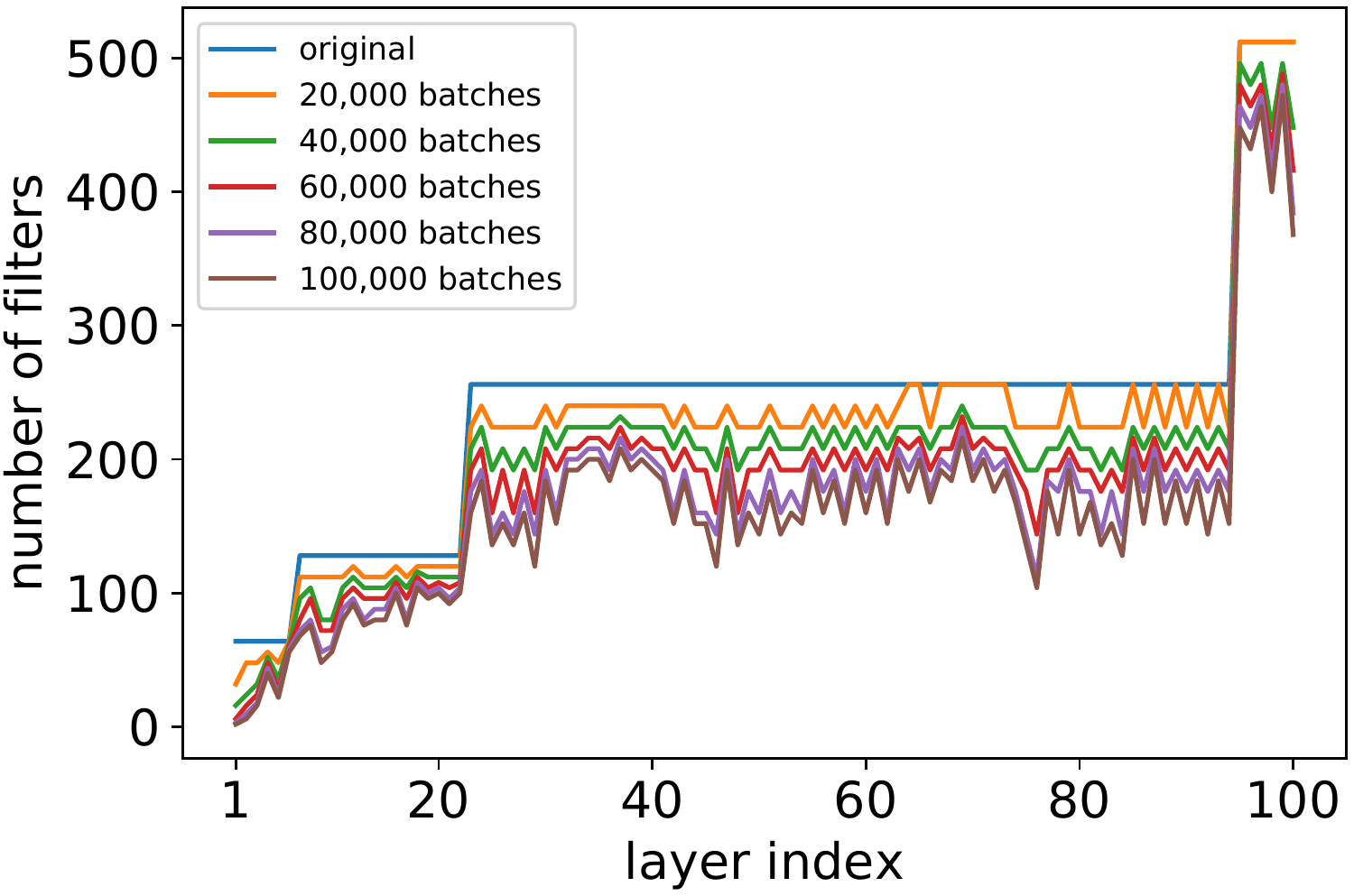} 
	\end{subfigure}
	\caption{Left: the remaining percentage of filters at the first layer in the four stages respectively. Right: the remaining width of all the pruned layers every 20,000 batches with a batch size of 128.}
	\label{fig-resnet152}
\end{figure}

Binary Filter Search enables not only the full use of the low-quality samples but also the adaptive pruning granularity. We present in Fig. \ref{fig-resnet152} the width of each layer of ResNet-152 (AOFP-E1). We pick the first layer in each of the four stages as the representatives, which originally have 64, 128, 256 and 512 filters, respectively. As AOFP proceeds, we show the remaining percentage of filters. Then we plot the remaining width of each layer every 20,000 batches. It can be observed that: \textbf{1)} AOFP automatically figures out that the first layer in stage2 can be pruned significantly, and chooses to prune it with large granularity (8 filters every time) at the beginning, then gradually reduces the granularity in order for more fine-grained pruning. However, AOFP always prunes 16 filters from the first layer in stage5. \textbf{2)} The adaptive granularity enables global progressive pruning, i.e., the reduction in the total FLOPs does not come from the extreme squeeze of several layers, nor pruning some at the beginning and others at the end. Instead, the network structure shrinks globally, steadily and progressively, which is more likely to result in high accuracy.

\subsection{AOFP for Destructive CNN Re-design}\label{sec-exp-4}
The above experiments are focused on pruning an existing mature CNN architecture for compression and acceleration. We then seek to use AOFP to re-design the CNN in order to reach a higher level of accuracy with the same computational budgets. To this end, we first train a scaled network from scratch and then apply AOFP until its FLOPs are reduced to the same level as the original model. In this way, we obtain a network where some layers are wider than the original architecture and some are narrower, so we call this process CNN Re-design. Concretely, we first scale the width of VGG and ResNet-50 (only the internal layers) by $1.5\times$ and $1.25\times$, respectively, and apply AOFP using the same hyper-parameters as Sect. \ref{sec-exp-2}. Though the pruned models outperform the baselines by a clear margin (Table. \ref{exp-table-redesign}), we still need to figure out whether the improvement is due to the better structure or the parameters initialized using the scaled model, so a counterpart with the same structure is trained from scratch, which delivers an accuracy higher than the baseline but lower than the pruned model. In this way, we safely claim the superiority of our optimized models over the tidy baselines. We present the discovered structures in the appendix to encourage further studies.
\begin{table}[t]
	\caption{Results of CNN Re-design by AOFP.}
	\label{exp-table-redesign}
	\begin{center}
		\begin{small}
			\begin{tabular}{llccccccc}
				\toprule
									& Top-1 	& 	FLOPs	& 	CPU 	& GPU		\\
				\midrule
				VGG base			&	93.38	&	313M	& 	343		&	6336	\\
				Scaled $1.5\times$			&	93.95	&	703M	&	205		&	3228	\\
				Re-design-pruned	&	94.03	&	312M	&	351		&	8099	\\
				\textbf{Re-design-scratch}	&	\textbf{93.69}	&	\textbf{312M}	&	-		&	-	\\
				\midrule
				Res50 base			&	75.34	&	3.85G	&	14.4	&	437		\\
				Scaled $1.25\times$				&	76.60	&	5.28G	&	11.2	&	353		\\
				Re-design-pruned	&	76.47	&	3.83G	&	14.2	&	430		\\
				\textbf{Re-design-scratch}	&	\textbf{76.30}	&	\textbf{3.83G}	&	-		&	-		\\
				\bottomrule
			\end{tabular}
		\end{small}
	\end{center}
\end{table}

\section{Conclusion}
We proposed Approximated Oracle Filter Pruning (AOFP), which features high quality of importance estimation, reasonable time complexity and no need for heuristic knowledge. We proposed a new CNN design paradigm, where we scale the network and apply AOFP to optimize its width to reach a higher level of accuracy without extra computational budgets, which can be used for refinement before a CNN is released. We empirically found out that the structural change in CNNs can be analyzed with local information only, which may inspire further theoretical researches.

\section*{Acknowledgement}
This work was supported by the National Key R\&D Program of China (No. 2018YFC0807500), National Natural Science Foundation of China (No. 61571269), National Postdoctoral Program for Innovative Talents (No. BX20180172), and the China Postdoctoral Science Foundation (No. 2018M640131). We sincerely thank the reviewers for their comments.

\bibliography{aofpbib}
\bibliographystyle{icml2019}

\section*{Appendix \\ Optimized Structures Discovered by AOFP}

\textbf{VGG on CIFAR-10.} 

In the re-designed VGG structure, each layer has 44, 80, 160, 180, 360, 360, 256, 224, 192, 56, 80, 192, 192 filters, respectively. Compared to the baseline, this model requires a roughly equal amount of FLOPs (312 v.s. 313 MFLOPs), but runs $1.27\times$ as fast (8099 examples/sec v.s. 6366 examples/sec), using CUDA9.2 and Tensorflow 1.10 on a GTX 1080Ti GPU. Interestingly, the wide layers at the early stages do not cause computational bottlenecks.

\textbf{ResNet-50 on ImageNet.}

\begin{figure}[b] 
	\includegraphics[width=0.8\columnwidth]{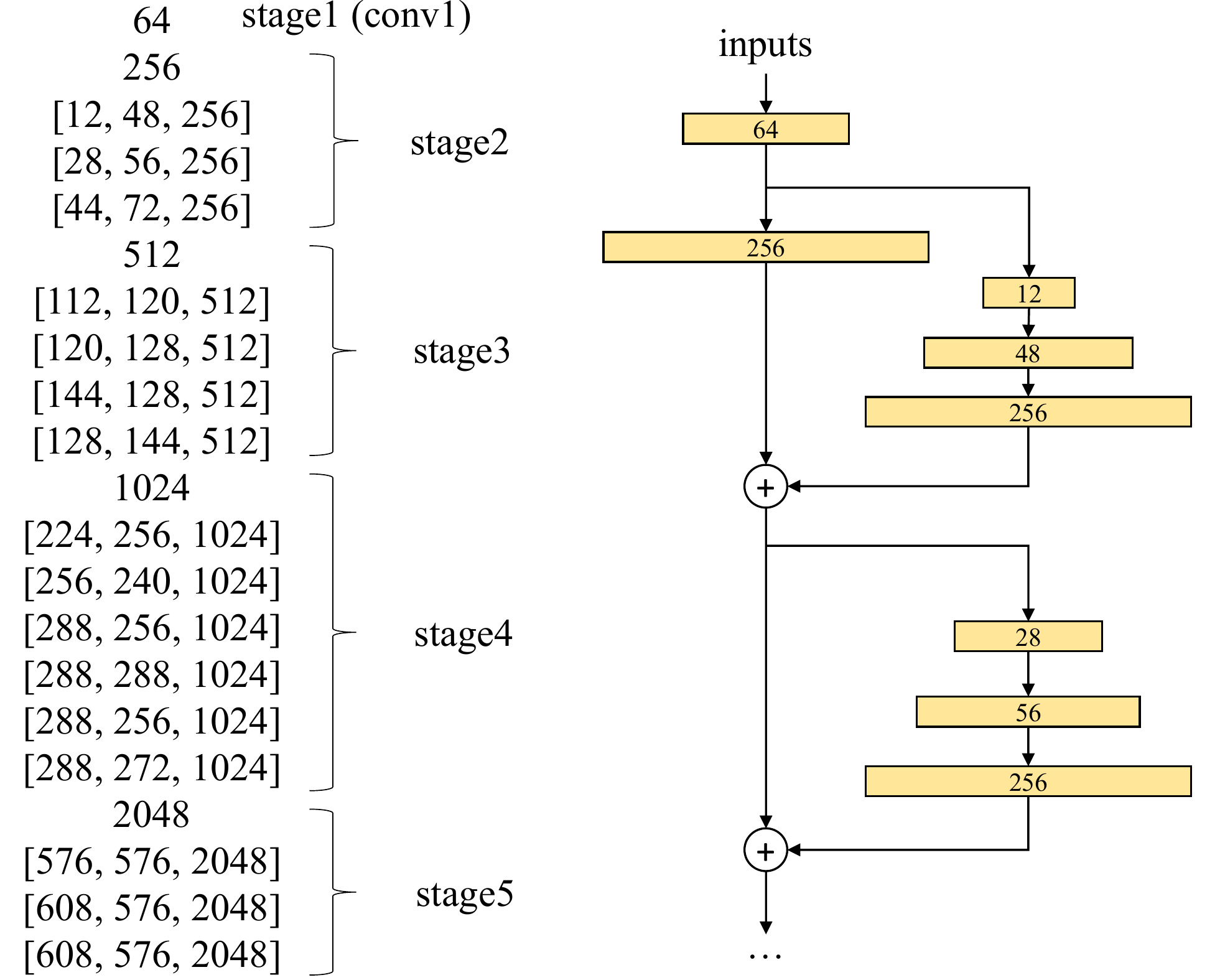}
	\caption{The structure of the re-designed ResNet-50. We present the number of filters of each convolutional on the left, and depict several layers at the beginning for example on the right.}
	\label{resnet50-sketch}
\end{figure}

The discovered structure of the re-designed ResNet-50 is depicted in Fig. \ref{resnet50-sketch}.

\begin{figure}[b] 
	\begin{subfigure}{0.49\columnwidth}\label{redesigned-width-plot-vc}
		\includegraphics[width=1\linewidth]{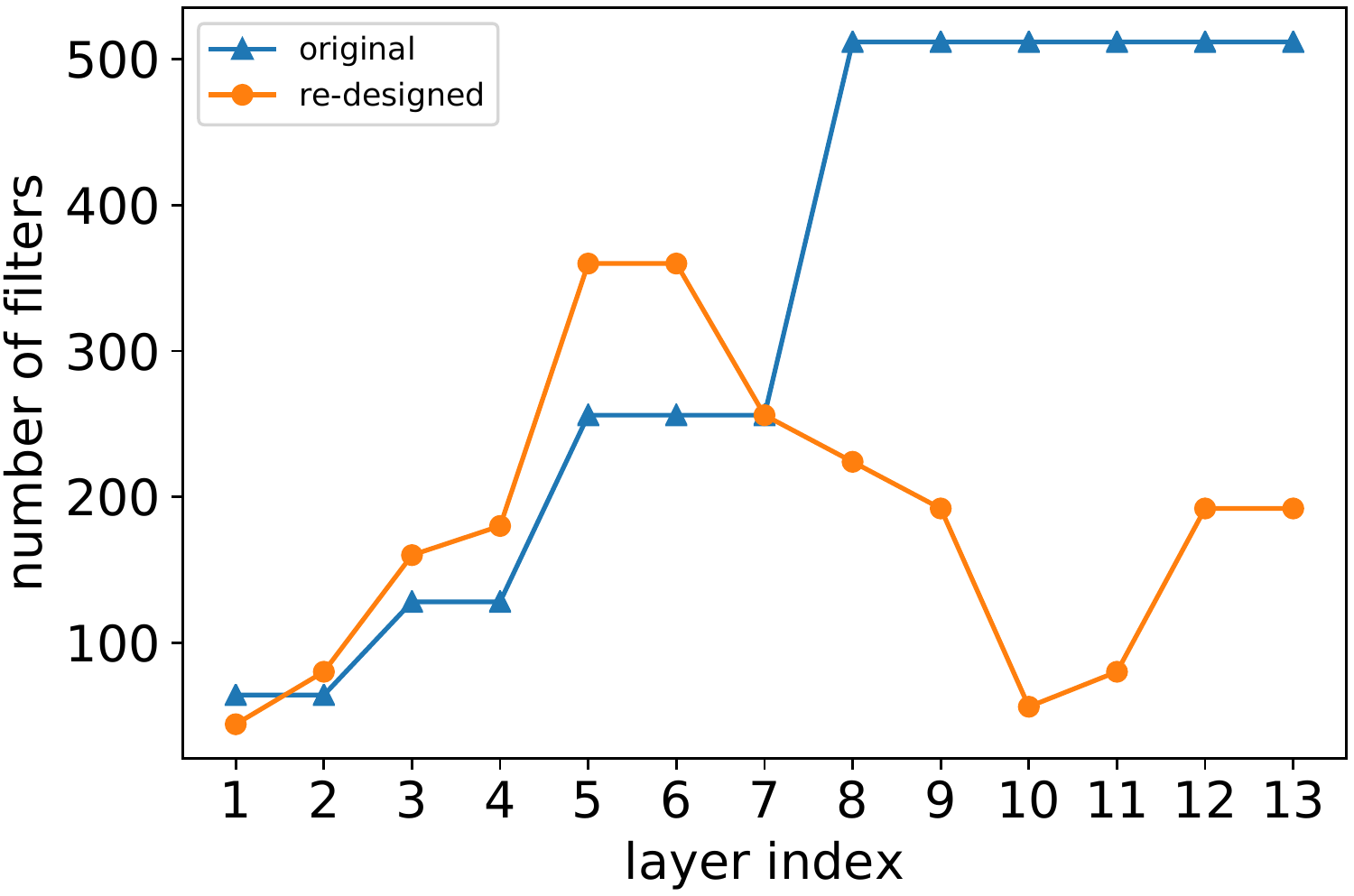} 
		\caption{VGG on CIFAR10.}
	\end{subfigure}
	\begin{subfigure}{0.49\columnwidth}\label{redesigned-width-plot-resnet50}
		\includegraphics[width=1\linewidth]{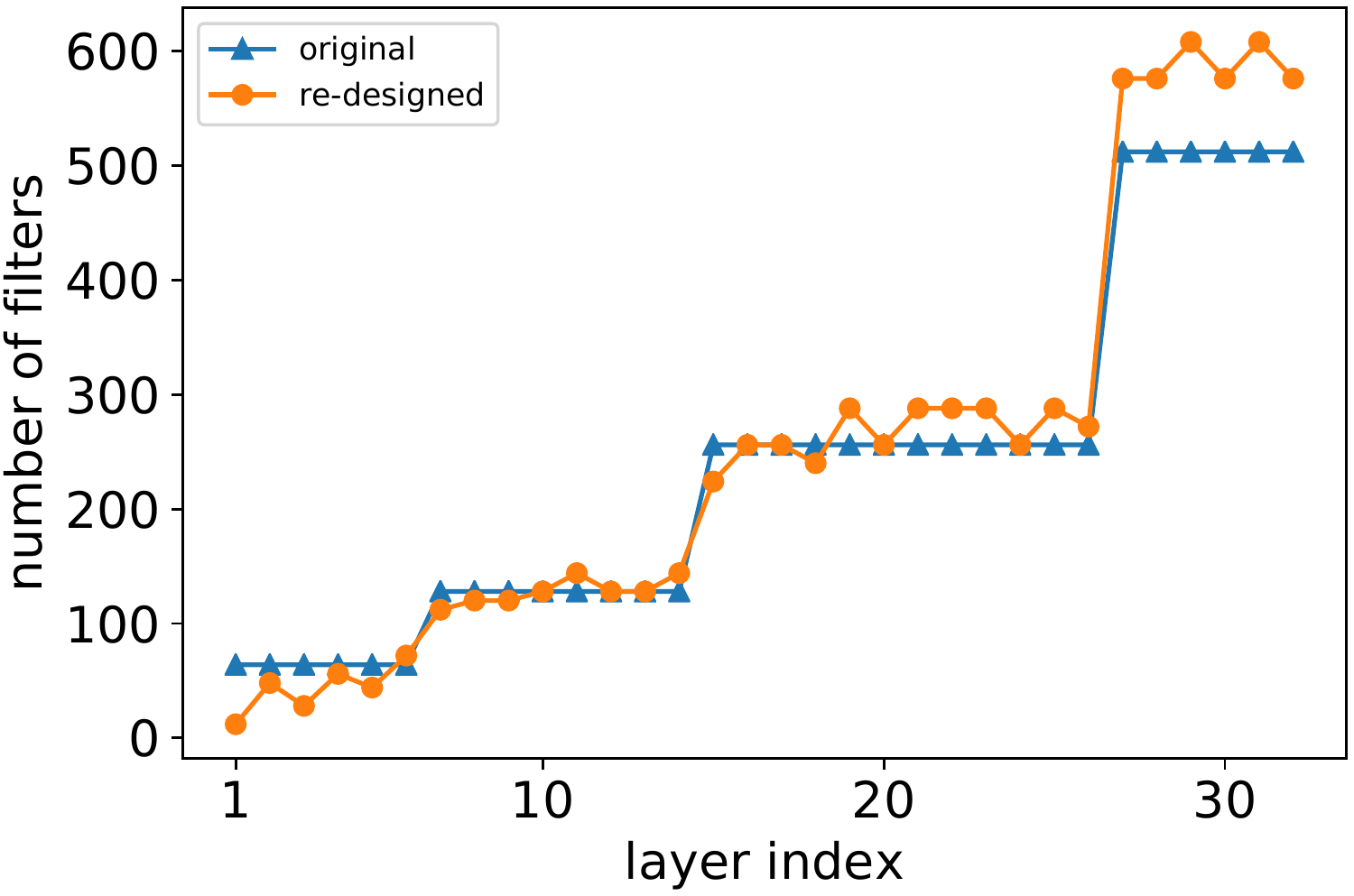} 
		\caption{ResNet-50 on ImageNet.}
	\end{subfigure}
	\vspace{-0.1in}
	\caption{Layer width of the re-designed models in comparison with the original. Note again that only the internal layers of ResNet-50 (i.e., the first two layers in each residual block) are shown.}
	\label{redesigned-width-plot}
\end{figure}

A comparison of the re-designed structures and the original models are presented in Fig. \ref{redesigned-width-plot}.

\end{document}